\documentclass[11pt]{article}

\usepackage[preprint]{acl}

\usepackage{times}
\usepackage{latexsym}

\usepackage[T1]{fontenc}

\usepackage[utf8]{inputenc}

\usepackage{microtype}

\usepackage{inconsolata}

\usepackage{graphicx}

\usepackage{setspace} 
\usepackage{refcount} 
\usepackage{upquote}  
\usepackage{booktabs}
\usepackage{subcaption}
\usepackage{multirow}
\usepackage{xcolor}
\usepackage{colortbl}
\usepackage{xurl}
\usepackage{hyperref}
\usepackage{amsmath}
\usepackage{array}
\usepackage{natbib}
\usepackage{amssymb}
\usepackage{tcolorbox}
\usepackage{float}
\usepackage{placeins}
\usepackage{pifont} 

\usepackage{tabularx}
\tcbuselibrary{breakable,skins}

\newcommand{\res}[2]{#1{\scriptsize $\pm$#2}}

\newtcolorbox{promptbox}[1]{%
    enhanced,
    breakable,
    colback=black!2,
    colframe=black!45,
    boxrule=0.5pt,
    arc=1mm,
    left=2mm,
    right=2mm,
    top=1.5mm,
    bottom=1.5mm,
    before skip=6pt,
    after skip=6pt,
    fonttitle=\bfseries\small,
    fontupper=\small\ttfamily,
    title={#1},
}

\title{
Stepwise Reasoning Verification: Don’t Generate, Classify!
}

\author{
 \textbf{Yifei Shi\textsuperscript{1}},
 \textbf{Michalis Korakakis\textsuperscript{1,2}},
 \textbf{Rami Aly\textsuperscript{1}},
 \textbf{Eric Chamoun\textsuperscript{1}}, \\
 \textbf{Lukasz Szpruch\textsuperscript{2,3}},
 \textbf{Andreas Vlachos\textsuperscript{1, *}}
\\
 \textsuperscript{1}University of Cambridge,
 \textsuperscript{2}The Alan Turing Institute,
 \textsuperscript{3}University of Edinburgh
\\
 \small{
  \textsuperscript{*}\textbf{Correspondence:} \href{mailto:av308@cam.ac.uk}{\texttt{av308@cam.ac.uk}}
 }
}

\begin{document}
\maketitle

\begin{abstract}
Verifying the reasoning of large language models is increasingly important as they are applied to complex mathematical and scientific problems. Existing work focuses on large decoder models and reports mixed results for generative and discriminative verifiers, leaving unclear which paradigm is more effective and whether large models are necessary. In this work, we cast stepwise verification as a natural language inference task: the problem and partial solution prefix form the premise, while the next reasoning step serves as the hypothesis to be classified as entailed or non-entailed. This formulation enables error localisation at individual reasoning steps and direct comparison of encoder classification, decoder classification, and decoder verdict generation across model scales and annotation budgets. Across five reasoning datasets, a 0.15B encoder classifier outperforms most generative verifiers with up to 4B parameters. On identical decoder backbones, classification outperforms generation from 0.5B to 4B parameters and matches it on the strongest 7B backbone---reaching comparable performance with roughly one tenth as many annotated problems. Our findings establish classification as a parameter- and data-efficient approach to stepwise verification and show that strong verification does not require large generative decoders.
\end{abstract}

\section{Introduction}
\label{sec:intro}

\begin{figure}
    \centering
    \includegraphics[width=1\linewidth]{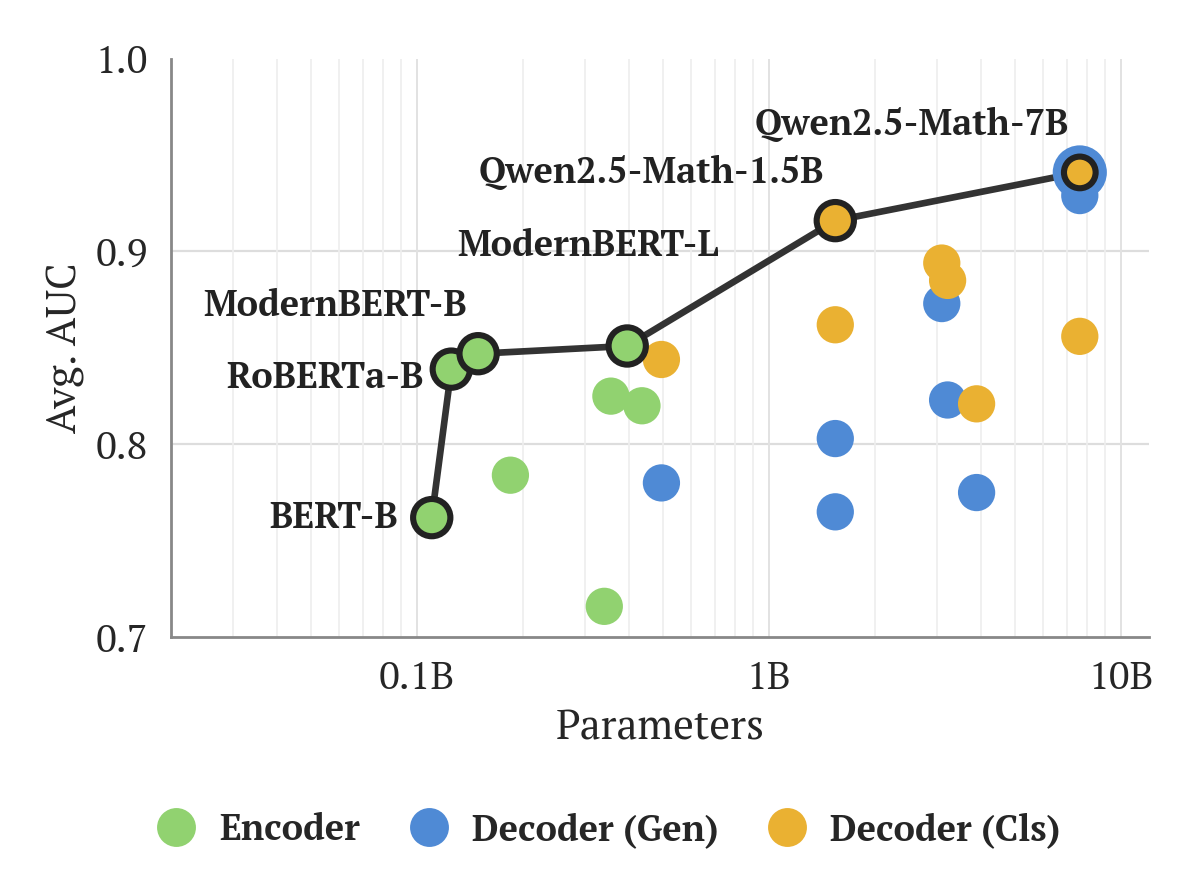}
    \caption{\textbf{Classifiers lie on the performance--parameter Pareto frontier.} Average AUC over five reasoning datasets versus parameter count for encoder classifiers, decoder classifiers, and generative verifiers. The black line marks the empirical Pareto frontier. On identical decoder backbones, classification achieves higher average AUC than verdict generation up to 4B parameters.}
    \label{fig:teaser}
\end{figure}

LLMs are increasingly used for complex mathematical and scientific problem solving, making reliable verification of their reasoning essential: seemingly coherent solutions can contain invalid intermediate steps, and a correct final answer does not guarantee sound reasoning~\citep{daheim2024stepwise,zheng2025processbench,li2024evaluating}. This need also extends to core educational applications of LLMs, such as tutoring and automated feedback, where corrective guidance requires locating errors in a student's reasoning rather than simply judging the solution incorrect~\citep{srivatsa2025llms, daheim2024stepwise}. Stepwise verification addresses these challenges by assessing individual reasoning steps in context to detect and localise errors~\citep{zheng2025processbench}.


Existing stepwise verifiers rely on large decoder models: discriminative approaches assign step-level scores or labels~\citep{lightman2023let, wang2024math}, while generative approaches produce verdicts~\citep{zhang2025generative, khalifa2025process, zhao2026genprm}. However, recent work finds no consistent performance advantage for either paradigm~\citep{lee2025rethinking}, leaving open which approach is preferable and whether large decoder models are necessary for strong verification. Resolving these questions is especially important for real-world deployment, where running a large verifier at every reasoning step can impose substantial latency and computational costs.

To address these questions, we formulate stepwise verification as natural language inference~(NLI)~\citep{dagan2005pascal, bowman2015large}. We treat the problem statement and solution prefix as the premise and the next reasoning step as the hypothesis, classified as entailed or non-entailed. This formulation extends encoder-based NLI classification, where encoders remain competitive with far larger LLMs on discriminative tasks \citep{elshabrawy2025enabling}, to stepwise verification and provides a common task definition for comparing encoder classifiers, decoder classifiers, and generative verifiers.

We evaluate our NLI formulation on five datasets covering mathematical, formal, and multi-domain reasoning. To the best of our knowledge, we are the first to demonstrate that encoder classifiers with only hundreds of millions of parameters can outperform generative verifiers with several billion parameters at detecting invalid reasoning steps~(Figure~\ref{fig:teaser}). On identical decoder backbones, classification outperforms generative verification from 0.5B to 4B and matches it on the strongest 7B backbone. In data-scaling experiments, classification often achieves comparable performance with roughly one tenth as many annotated problems as generative verification. Collectively, our findings suggest that how a model is trained to verify stepwise reasoning matters more than its size, supporting the use of small classifiers in applications where repeated step-level verification must fit within limited time and compute budgets.

\section{Related Work}
\label{sec:related}

\paragraph{Verification objectives.}
\label{para:ver_paradigm_related}
Verification requires accurate error detection at a computational cost that permits repeated checks throughout a solution, making small, effective verifiers a practical priority. However, existing approaches typically rely on large decoder backbones, using dedicated scoring heads~\citep{lightman2023let, wang2024math, zhang2025lessons}, generating verdicts through next-token prediction~\citep{zhang2025generative}, or generating verification rationales before predicting a verdict~\citep{khalifa2025process, zhao2026genprm}. Whether this scale is necessary depends in part on how the model is trained to verify. \citet{lee2025rethinking} find that the performance of classification and generation varies across verification settings, while differences in model size, supervision, and inference computation further complicate comparisons. We therefore compare classification and direct-verdict generation on identical decoder backbones and training instances across model sizes and annotation budgets, testing whether classification can reduce the model size and supervision needed for accurate verification. Appendix~\ref{appendix:literature} compares these studies by the factors each holds fixed.


\paragraph{Verification as inference.}
\label{para:nli_related}
Assessing a reasoning step requires relating it to the problem and preceding steps. NLI provides a formulation of this task by assessing whether a hypothesis follows from a premise~\citep{dagan2005pascal}: the problem and preceding reasoning form the premise, and the current step is the hypothesis. Prior work uses entailment and deductive judgements to evaluate intermediate reasoning~\citep{zhu2024deductive, ling2023deductive, zhou2024don, sadeddine2025verifying}, including existing NLI models that detect unsupported or contradictory claims within reasoning chains~\citep{golovneva2022roscoe, prasad2023receval}. Related approaches examine mathematical premise selection~\citep{ferreira2020natural}, verify steps through autoformalisation and theorem proving~\citep{hu2025stepproof, sadeddine2025verifying}, or extract validity signals from attention structure~\citep{noel2026geometry}.  We build on inference-based evaluation by training classifiers directly on step-level error labels. The premise--hypothesis formulation provides a common input for encoder and decoder verifiers, allowing us to compare architectures and training objectives on identical labelled instances.

\paragraph{Small encoder verifiers.}
Encoders are a well-motivated alternative to large decoder verifiers given their established use in NLI and their effectiveness and parameter efficiency on discriminative language-understanding and reasoning tasks~\citep{weller2026seq, roy2025causal, nielsen2025encoder, popovivc2025extractive, elshabrawy2025enabling}. Their potential is also supported by \citet{gu2023don}, who show that scoring candidate plans with BERT-base can outperform generative approaches while requiring less supervision. In reasoning verification, DIVERSE~\citep{li2023making} fine-tunes a DeBERTa encoder with solution-level supervision and auxiliary step labels, demonstrating that encoders can learn step-level predictions. Its evaluation, however, focuses on improving final-answer selection. Conversely, we examine the effectiveness of encoders as standalone step-level verifiers, comparing them with decoder classifiers and generative verifiers across datasets, model sizes, and annotation budgets. This directly tests whether small classifiers can match larger generative models at detecting reasoning errors, and how much supervision they require.

\section{Methodology}
\label{sec:method}

\subsection{Stepwise Verification as Natural Language Inference}
\label{subsec:task_form}

Let a reasoning trajectory consist of a problem statement $Q$ and an ordered sequence of steps $s_1,\ldots,s_n$. To verify each step $s_i$, we construct a premise containing the problem and all preceding steps, with the current step as the hypothesis:

\begin{equation}
    \begin{aligned}
        P_i &= Q \oplus s_1 \oplus \cdots \oplus s_{i-1},\\
        H_i &= s_i,
    \end{aligned}
    \label{eq:premise_hyp}
\end{equation}
where $\oplus$ denotes concatenation. Each verification instance is $(x_i,y_i)$, with
\[
    \begin{aligned}
        x_i &= (P_i,H_i),\\
        y_i &\in \{\textit{entailment},\textit{non-entailment}\}.
    \end{aligned}
\]

These labels represent step validity in context: $H_i$ is labelled \textit{entailment} if it is supported by $P_i$, and \textit{non-entailment} otherwise. For example, given the problem ``Solve $2x+3=7$'' and the prefix ``$2x=4$,'' the next step ``$x=2$'' is entailed, whereas ``$x=4$'' is non-entailed. This formulation enables precise localisation of intermediate reasoning errors and provides a common task definition for encoder classification, decoder classification, and generative verification.

Concretely, we use the predicted probability of \textit{non-entailment} as the step-level error score:
\begin{equation}
    p_i = p_\theta\bigl(\textit{non-entailment}\mid x_i\bigr).
    \label{eq:non_entailment_score}
\end{equation}

For a trajectory whose first annotated error occurs at step $k$, we retain the $k$ instances $\{(x_i,y_i)\}_{i=1}^{k}$ and assign
\[
    y_i=
    \begin{cases}
        \textit{entailment}, & i<k,\\
        \textit{non-entailment}, & i=k.
    \end{cases}
\]

We discard subsequent steps because their premises already contain an annotated error. For fully correct trajectories, we retain all $n$ steps and label each as \textit{entailment}. Each retained instance is scored separately using its full premise $P_i$.

\subsection{Classification and Generative Verification}
\label{subsec:paradigm}

\begin{figure}[t!]
    \centering
    \includegraphics[width=1\linewidth]{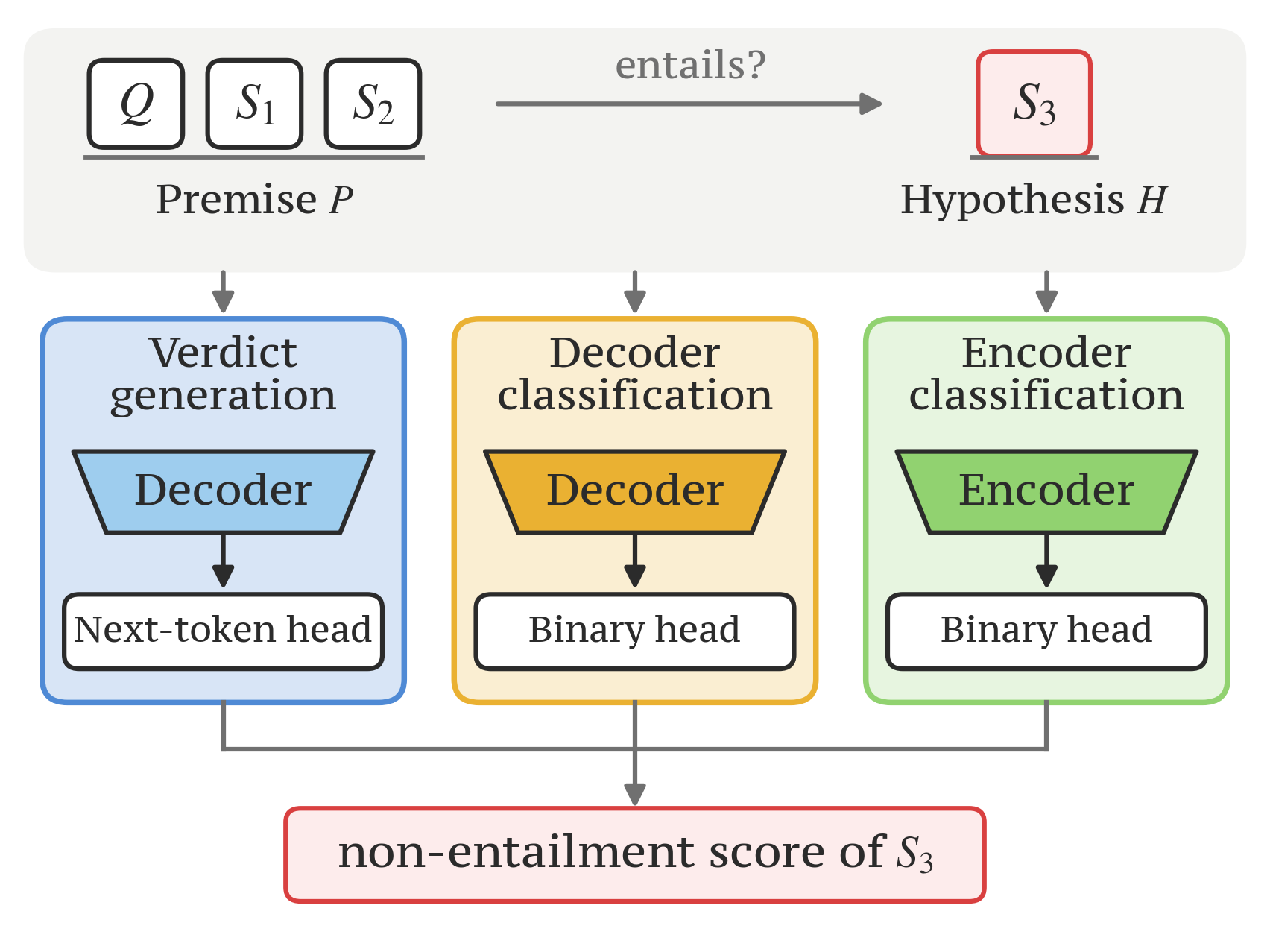}
    \caption{
    \textbf{Stepwise reasoning verification as natural language inference.} To verify reasoning step $S_3$, the premise $P$ combines the problem statement $Q$ with preceding steps $S_1$ and $S_2$, while the hypothesis $H$ is $S_3$. All three verifiers assign a \textit{non-entailment} score to this premise--hypothesis pair.
    }
    \label{fig:formulation}
\end{figure}

To test whether strong verification requires large decoder models, we compare encoder classification, decoder classification, and decoder verdict generation across model scales, using identical premise--hypothesis instances (Figure~\ref{fig:formulation}).

\begin{enumerate}
    \item \textbf{Encoder classification.}
    An encoder maps the premise--hypothesis pair $x_i$ to a representation $h_i^{\mathrm{enc}}$. A binary classification head then produces the non-entailment score $p_i$.

    \item \textbf{Decoder verdict generation.}
    We fine-tune the decoder to generate a single verdict token: \texttt{Yes} for entailment or \texttt{No} for non-entailment. At inference, we compute the non-entailment score by normalising over the two verdict tokens:
    \begin{equation}
        p_i =
        \frac{\exp z_i^{\mathrm{No}}}
        {\exp z_i^{\mathrm{Yes}}+\exp z_i^{\mathrm{No}}},
        \label{eq:verdict_score}
    \end{equation}
    where $z_i^{\mathrm{Yes}}$ and $z_i^{\mathrm{No}}$ are their logits at the last prompt position.

    \item \textbf{Decoder classification.}
    The same decoder processes $x_i$ causally. We take its hidden state $h_i^{\mathrm{dec}}$ at the final non-padding input token and apply a linear head with two outputs:
    \[
    \begin{aligned}
        z_i^{\mathrm{Yes}}
            &= w_{\mathrm{Yes}}^\top h_i^{\mathrm{dec}}
               + b_{\mathrm{Yes}},\\
        z_i^{\mathrm{No}}
            &= w_{\mathrm{No}}^\top h_i^{\mathrm{dec}}
               + b_{\mathrm{No}}.
    \end{aligned}
    \]
    The non-entailment score uses the same normalisation as Equation~\ref{eq:verdict_score}. We initialise $w_{\mathrm{Yes}}$ and $w_{\mathrm{No}}$ from the language-model output embeddings of the two verdict tokens and set both biases to zero.
\end{enumerate}

Both decoder variants use supervised fine-tuning (SFT); we refer to them as \textbf{decoder generation SFT} and \textbf{decoder classification SFT}. They share the pretrained backbone, input prompt, and training instances, but differ in their output heads and training objectives. Although both make the same binary verification decision, generation minimises next-token cross-entropy over the full vocabulary, whereas classification minimises cross-entropy over the two verification labels. Comparisons between encoders and decoders additionally reflect differences in architecture, pretraining, and context capacity.


\section{Experiment Setup}
\label{sec:exp}


\subsection{Models and Datasets}
\label{subsec:models_datasets}

Our NLI formulation motivates evaluating encoder models widely used for this task: BERT~\citep{devlin2019bert}, RoBERTa~\citep{liu2019roberta}, DeBERTa~\citep{he2023debertav3}, and ModernBERT~\citep{warner2025smarter}, each at \textit{base} and \textit{large} scales. For decoders, we use pretrained checkpoints from Qwen2.5~\citep{yang2024qwen2} at 0.5B, 1.5B, 3B, and 7B; Qwen2.5-Math~\citep{yang2024qwen2math} at 1.5B and 7B; Llama-3.2-3B~\citep{grattafiori2024llama}; and Gemma-3-4B~\citep{team2025gemma}. Each decoder is fine-tuned separately for verdict generation and classification, allowing us to compare the two objectives on the same pretrained backbone.

We use five English datasets common in stepwise verification~\citep{zhang2025lessons, khalifa2025process, lee2025rethinking}, spanning mathematical, formal, and multi-domain reasoning: PRM800K~\citep{lightman2023let}, MathShepherd~\citep{wang2024math} (MS), Epic50K~\citep{sun2025efficient}, FoVer~\citep{kamoi2026efficient}, and MMLU-Pro-CoT~\citep{zeng2025versaprm} (MMLU). Their step labels come from different sources (Table~\ref{tab:datasets_brief}) and measure different things: human judgements (PRM800K) and formal verifiers (FoVer) assess whether a step is valid, whereas rollout labels (MathShepherd) assess whether the prefix can still reach the correct answer, which is closer to the step's utility than to its validity~\citep{lee2025evaluating}.

\begin{table}[t!]
    \centering
    \small
    \setlength{\tabcolsep}{2.6pt}
    \begin{tabular}{@{}lllrrr@{}}
        \toprule
        \textbf{Dataset} & \textbf{Domain} & \textbf{Labels}
        & \textbf{Train} & \textbf{Test} & \textbf{Err.} \\
        \midrule
        PRM800K      & Math   & Human        & 46.0k & 70.5k & 8.3 \\
        MS & Math   & Rollout      & 37.0k & 80.9k & 17.5 \\
        Epic50K      & Math   & Contrib.     & 42.7k & 37.8k & 6.0 \\
        FoVer        & Formal & Verifier     & 34.3k & 10.8k & 24.0 \\
        MMLU     & Multi  & Synthetic    & 38.0k & 49.5k & 9.6 \\
        \bottomrule
    \end{tabular}
    \caption{\textbf{The five datasets differ in domain, annotation source, and class imbalance.} \textit{Labels} is the source of the step annotations: human judgement, rollout-based answer reachability, contribution-based automatic annotation, formal verification tools, or synthetic annotation. \textit{Train} counts training problems, \textit{Test} counts premise--hypothesis instances, and \textit{Err.} is the percentage of non-entailment instances in the test split. Full statistics are in Table~\ref{tab:dataset_statistics}.}
    \label{tab:datasets_brief}
\end{table}

We construct all splits at the question level, keeping solutions to the same question together to prevent overlap between training and evaluation. For PRM800K and MathShepherd, we retain the released test partitions and exclude their questions from training. For the remaining datasets, we hold out 10\% of questions for testing. After converting solutions into premise--hypothesis instances (Section~\ref{subsec:task_form}), we retain one copy of each duplicate pair with matching labels and discard pairs assigned conflicting labels.

To keep the numbers of training problems comparable across datasets, we cap PRM800K, MathShepherd, and MMLU-Pro-CoT at between 37k and 46k problems per dataset and retain all available training problems from Epic50K and FoVer. These sets define our \textit{Full} training condition.


Training splits are balanced by undersampling within each step position, so that position does not predict the label; validation and test splits keep their natural label distribution. Inputs beyond a model's context limit are truncated from the left, keeping the step and its nearest context; this mainly affects Epic50K, where a third of test instances exceed 512 tokens (Appendix~\ref{appendix:datasets}).

\subsection{Training Setup}
\label{subsec:training_setup}

All models are trained with \textbf{full-parameter fine-tuning} using the 8-bit AdamW optimiser \citep{dettmers2022optimizers}. We repeat each SFT experiment with three random seeds and report results aggregated across the three runs. Full implementation details are provided in Appendix~\ref{appendix:impl_details}.

\subsection{Metrics}
\label{subsec:metrics}

Following prior work~\citep{xia2025evaluating, kamoi2026efficient}, we report step-level area under the receiver operating characteristic curve~(AUC), computed over separately scored premise--hypothesis instances. AUC measures discrimination between the two labels without requiring a decision threshold and does not directly depend on the class proportions of the test splits, which retain their natural label distributions.


\section{Results \& Analysis}
\label{sec:results}

\subsection{Main Results}
\label{subsec:ver_results}

\par Table~\ref{tab:auc_results} shows that the training objective matters more than model size. On identical decoder backbones up to 4B, classification achieves an average AUC of $0.870$, compared with $0.803$ for generation---a gain of $0.067$ without increasing backbone size. Encoder classifiers with 0.11B--0.44B parameters average $0.806$ AUC, comparable to these generative decoders despite having roughly one ninth as many parameters on average. Qwen2.5-0.5B trained for classification ($0.844$ AUC) also outperforms every generative decoder up to 4B except Qwen2.5-3B. At 7B, classification matches generation on Qwen2.5-Math, demonstrating that it remains competitive on larger backbones. We next examine these results in more detail, focusing first on the training objective and then on model scale.



\begin{table*}[t]
    \centering
    \small
    \setlength{\tabcolsep}{5pt}
    
    \begin{tabular}{l c *{5}{c} c c}
    \toprule
    \textbf{Model}
    & \textbf{Paradigm}
    & \textbf{PRM}
    & \textbf{MS}
    & \textbf{Epic}
    & \textbf{FoVer}
    & \textbf{MMLU}
    & \textbf{Avg.}
    & $\boldsymbol{\Delta}$ \\
    \midrule
    
    \multicolumn{9}{l}{\textit{Decoders}} \\
    
    \multirow{2}{*}{Qwen2.5-0.5B}
    & Gen.
    & \res{.680}{.011}
    & \res{.665}{.010}
    & \res{.968}{.001}
    & \res{.764}{.016}
    & \res{.826}{.004}
    & \res{.780}{.008}
    & \multirow{2}{*}{+.064} \\
    
    & Cls.
    & \res{.745}{.097}
    & \res{.769}{.010}
    & \res{.974}{.002}
    & \res{.894}{.046}
    & \res{.838}{.004}
    & \res{.844}{.032}
    & \\
    
    \addlinespace[3pt]
    
    \multirow{2}{*}{Qwen2.5-1.5B}
    & Gen.
    & \res{.707}{.040}
    & \res{.589}{.012}
    & \res{.969}{.001}
    & \res{.701}{.026}
    & \res{.858}{.008}
    & \res{.765}{.017}
    & \multirow{2}{*}{+.097} \\
    
    & Cls.
    & \res{.780}{.070}
    & \res{.824}{.003}
    & \res{.978}{.002}
    & \res{.860}{.158}
    & \res{.867}{.004}
    & \res{.862}{.047}
    & \\
    
    \addlinespace[3pt]
    
    \multirow{2}{*}{Qwen2.5-Math-1.5B}
    & Gen.
    & \res{.767}{.013}
    & \res{.617}{.012}
    & \res{.973}{.001}
    & \res{.792}{.001}
    & \res{.864}{.005}
    & \res{.803}{.007}
    & \multirow{2}{*}{\textbf{+.113}} \\
    
    & Cls.
    & \res{.838}{.068}
    & \res{.890}{.002}
    & \res{.985}{.000}
    & \res{.974}{.001}
    & \res{.891}{.001}
    & \res{.916}{.014}
    & \\
    
    \addlinespace[3pt]
    
    \multirow{2}{*}{Qwen2.5-3B}
    & Gen.
    & \res{.843}{.004}
    & \res{.790}{.010}
    & \res{.976}{.000}
    & \res{.868}{.071}
    & \res{.886}{.003}
    & \res{.873}{.018}
    & \multirow{2}{*}{+.021} \\
    
    & Cls.
    & \res{.858}{.004}
    & \res{.808}{.097}
    & \res{.977}{.008}
    & \res{.977}{.001}
    & \res{.849}{.085}
    & \res{.894}{.039}
    & \\
    
    \addlinespace[3pt]
    
    \multirow{2}{*}{Llama-3.2-3B}
    & Gen.
    & \res{.782}{.010}
    & \res{.613}{.009}
    & \res{.970}{.001}
    & \res{.894}{.011}
    & \res{.855}{.004}
    & \res{.823}{.007}
    & \multirow{2}{*}{+.062} \\
    
    & Cls.
    & \res{.811}{.001}
    & \res{.795}{.003}
    & \res{.977}{.000}
    & \res{.965}{.004}
    & \res{.875}{.002}
    & \res{.885}{.002}
    & \\
    
    \addlinespace[3pt]
    
    \multirow{2}{*}{Gemma-3-4B}
    & Gen.
    & \res{.689}{.093}
    & \res{.598}{.014}
    & \res{.967}{.003}
    & \res{.771}{.036}
    & \res{.850}{.002}
    & \res{.775}{.030}
    & \multirow{2}{*}{+.046} \\
    
    & Cls.
    & \res{.761}{.041}
    & \res{.763}{.056}
    & \res{.972}{.004}
    & \res{.766}{.097}
    & \res{.843}{.008}
    & \res{.821}{.041}
    & \\
    
    \addlinespace[3pt]
    
    \multirow{2}{*}{Qwen2.5-7B}
    & Gen.
    & \res{.876}{.006}
    & \res{.887}{.002}
    & \res{.986}{.001}
    & \res{.985}{.001}
    & \res{.914}{.001}
    & \res{.929}{.002}
    & \multirow{2}{*}{-.073} \\
    
    & Cls.
    & \res{.856}{.003}
    & \res{.882}{.002}
    & \res{.922}{.095}
    & \res{.811}{.150}
    & \res{.811}{.089}
    & \res{.856}{.068}
    & \\
    
    \addlinespace[3pt]
    
    \multirow{2}{*}{Qwen2.5-Math-7B}
    & Gen.
    & \res{\textbf{.907}}{.001}
    & \res{.905}{.002}
    & \res{\textbf{.989}}{.001}
    & \res{\textbf{.990}}{.001}
    & \res{\textbf{.917}}{.002}
    & \res{\textbf{.941}}{.001}
    & \multirow{2}{*}{.000} \\
    
    & Cls.
    & \res{.899}{.001}
    & \res{\textbf{.910}}{.001}
    & \res{.988}{.000}
    & \res{.989}{.001}
    & \res{\textbf{.917}}{.001}
    & \res{\textbf{.941}}{.001}
    & \\
    
    \midrule
    
    \multicolumn{9}{l}{\textit{Encoders}} \\
    
    BERT-B (0.11B)
    & Cls.
    & \res{.743}{.004}
    & \res{.731}{.003}
    & \res{.765}{.001}
    & \res{.800}{.004}
    & \res{.770}{.005}
    & \res{.762}{.003}
    & \\
    
    BERT-L (0.34B)
    & Cls.
    & \res{.698}{.060}
    & \res{.705}{.015}
    & \res{.683}{.130}
    & \res{.720}{.088}
    & \res{.774}{.008}
    & \res{.716}{.060}
    & \\
    
    \addlinespace[2pt]
    
    RoBERTa-B (0.13B)
    & Cls.
    & \res{.777}{.007}
    & \res{.720}{.008}
    & \res{.972}{.001}
    & \res{.916}{.001}
    & \res{.812}{.002}
    & \res{.839}{.004}
    & \\
    
    RoBERTa-L (0.36B)
    & Cls.
    & \res{.755}{.042}
    & \res{.639}{.121}
    & \res{.972}{.001}
    & \res{.939}{.003}
    & \res{.822}{.010}
    & \res{.825}{.035}
    & \\
    
    \addlinespace[2pt]
    
    DeBERTa-B (0.18B)
    & Cls.
    & \res{.778}{.004}
    & \res{.617}{.073}
    & \res{.809}{.007}
    & \res{.907}{.005}
    & \res{.806}{.018}
    & \res{.784}{.021}
    & \\
    
    DeBERTa-L (0.44B)
    & Cls.
    & \res{.793}{.004}
    & \res{.700}{.004}
    & \res{.813}{.012}
    & \res{.941}{.008}
    & \res{.850}{.004}
    & \res{.820}{.007}
    & \\
    
    \addlinespace[2pt]
    
    ModernBERT-B (0.15B)
    & Cls.
    & \res{.780}{.011}
    & \res{.730}{.003}
    & \res{.973}{.001}
    & \res{.946}{.007}
    & \res{.804}{.005}
    & \res{.847}{.005}
    & \\
    
    ModernBERT-L (0.4B)
    & Cls.
    & \res{.783}{.011}
    & \res{.697}{.024}
    & \res{.976}{.001}
    & \res{.960}{.003}
    & \res{.839}{.004}
    & \res{.851}{.009}
    & \\
    
    
    
    
    
    




    
    \bottomrule
    \end{tabular}
    
    \caption{
    \textbf{Step-level verification across paradigms, architectures, and model scales.}
    We report step-level AUC as mean $\pm$ standard deviation over three random seeds.
    Bold marks the best result in each column among trained models.
    $\Delta$ denotes the difference in average AUC,
    $\mathrm{AUC}_{\mathrm{Cls}}-\mathrm{AUC}_{\mathrm{Gen}}$.
    \textit{Avg.}\ is the macro-average AUC over the five datasets.
    Encoder classifiers are reported in a separate block.
    }
    \label{tab:auc_results}
\end{table*}

\begin{figure}[t!]
    \centering
    \includegraphics[width=1\linewidth]{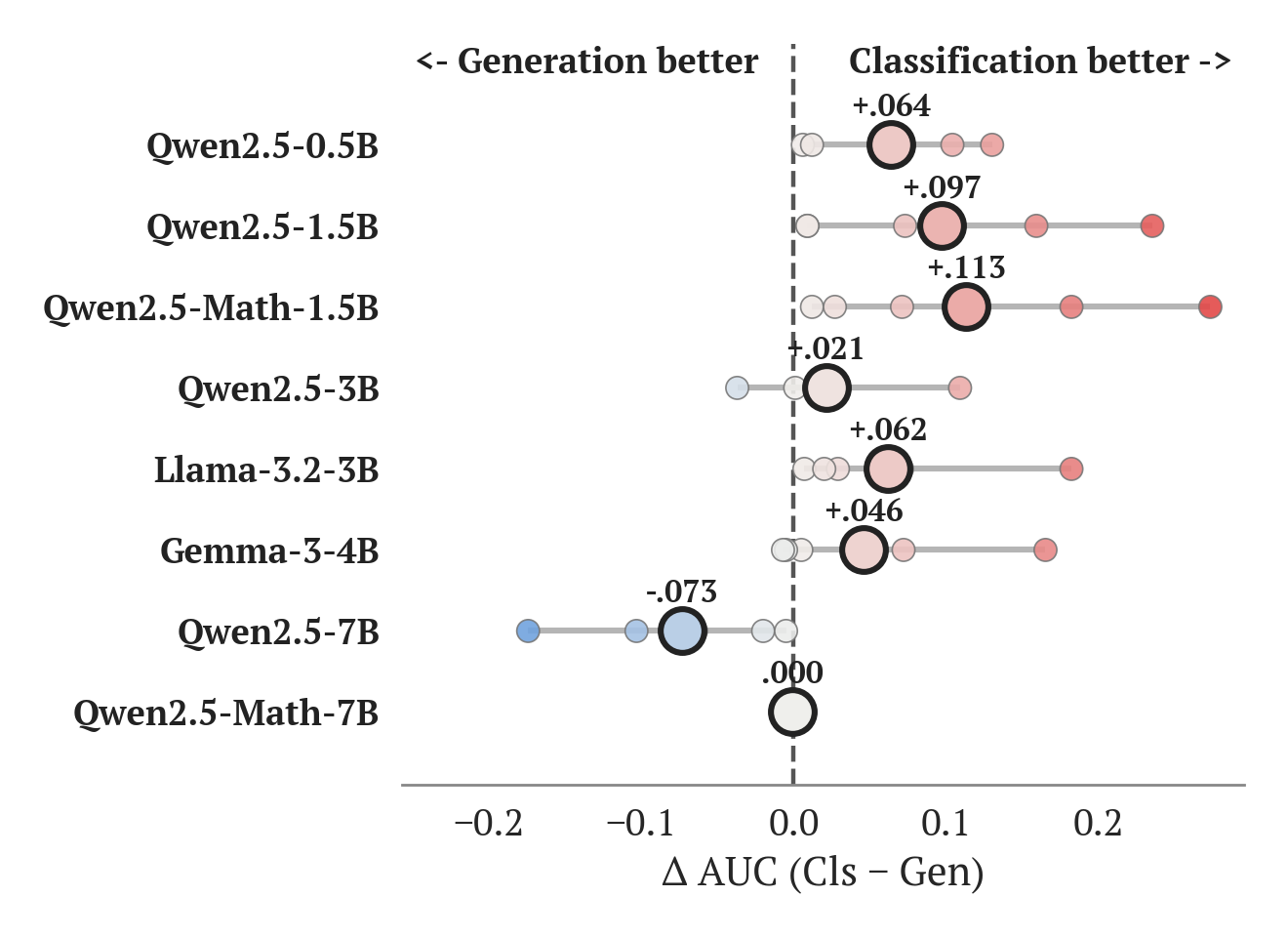}
    \caption{\textbf{Classification's advantage is largest at small scale and is gone or reversed at 7B.} $\Delta$ AUC (classification $-$ generation) per decoder backbone: small points are per-dataset deltas, large points are the mean. All five datasets favour classification at 0.5B--1.5B; classification matches or exceeds generation on average at seven of the eight backbones.}
    \label{fig:delta}
\end{figure}

\paragraph{Classification vs.\ generation.}
Figure~\ref{fig:delta} compares classification and generation on identical backbones. Classification outperforms generation on every dataset for all 0.5B and 1.5B backbones, with average gains of $0.06$--$0.11$ AUC. This advantage persists at 3B and 4B, with average gains of $0.02$--$0.06$ AUC on Qwen2.5-3B, Llama-3.2-3B, and Gemma-3-4B. At 7B, classification matches generation on Qwen2.5-Math, with both achieving $0.941$ average AUC. The general Qwen2.5-7B results suggest that more stable optimisation could further improve classification, as instability in some runs contributes to the $0.073$ average AUC gap relative to generation. Across datasets, classification delivers its largest gains on MathShepherd and FoVer: averaged over the six backbones up to 4B, it improves AUC by $0.16$ and $0.11$, respectively, compared with gains below $0.01$ on Epic50K and MMLU-Pro-CoT.


\paragraph{Model scale.}
Strong verification does not require large decoders. ModernBERT-Base~(ModernBERT-B), with only 0.15B parameters, achieves $0.847$ average AUC, outperforming every generative decoder up to 4B except Qwen2.5-3B, which has twenty times as many parameters. Increasing encoder size does not consistently improve performance (Table~\ref{tab:auc_results}). Domain specialisation, however, can improve verification without increasing model size: Qwen2.5-Math-1.5B outperforms its general-purpose counterpart on every dataset under both objectives.

\subsection{Head Initialisation Ablation}
\label{subsec:head_init}

\begin{table}[t!]
    \centering
    \small
    \setlength{\tabcolsep}{4pt}
    \renewcommand{\arraystretch}{1.08}
    \begin{tabular}{@{}lccc}
        \toprule
        \textbf{Dataset}
        & \textbf{Zero}
        & \textbf{Verbalizer} \\
        \midrule
        PRM800K
        & \res{.722}{.109}
        & \res{\textbf{.858}}{\textbf{.004}} \\
        MathShepherd
        & \res{.789}{.118}
        & \res{\textbf{.808}}{\textbf{.097}} \\
        Epic50K
        & \res{.909}{.125}
        & \res{\textbf{.977}}{\textbf{.008}} \\
        FoVer
        & \res{.857}{.130}
        & \res{\textbf{.977}}{\textbf{.001}} \\
        MMLU-Pro-CoT
        & \res{\textbf{.887}}{\textbf{.010}}
        & \res{.849}{.085} \\
        \midrule
        Avg.
        & \res{.833}{.098}
        & \res{\textbf{.894}}{\textbf{.039}} \\
        \bottomrule
    \end{tabular}
    \caption{\textbf{Verbalizer initialisation makes decoder classification stronger and more consistent across seeds.} Step-level AUC of Qwen2.5-3B classification, mean $\pm$ standard deviation over three seeds, with a zero-initialised head and with the head initialised from the verdict tokens, on the same data and seeds. Bold marks the higher mean and the lower standard deviation per row.}
    \label{tab:head_init}
\end{table}

\par Decoder classification introduces a task-specific head whose initialisation can affect performance and training stability. We compare two strategies: \textit{zero} initialisation sets $w_{\mathrm{Yes}}$, $w_{\mathrm{No}}$, and both biases to zero, initially assigning equal logits to both classes; \textit{verbalizer} initialisation sets the class weight vectors to the language-model output embeddings of \texttt{Yes} and \texttt{No}, reusing the pretrained weights for scoring these verdict tokens. We train Qwen2.5-3B under both initialisations using the same data and random seeds.

\par Table~\ref{tab:head_init} shows that verbalizer initialisation improves both performance and training stability, raising average AUC from $0.833$ to $0.894$ and reducing the standard deviation across seeds from $0.098$ to $0.039$. With zero initialisation, some runs substantially underperform the others. Verbalizer initialisation produces consistent results on PRM800K, Epic50K, and FoVer, with standard deviations of at most $0.008$. Overall, verbalizer initialisation improves average AUC by $0.061$ and reduces the standard deviation across seeds by approximately $60\%$, even with one underperforming run among the three seeds on MathShepherd and MMLU-Pro-CoT.

\subsection{Annotated Data Scaling Ablation}
\label{subsec:data_results}

\par To compare the data efficiency of classification and generation for step-level verification, we vary the number of annotated training problems over doubling budgets and include the full training set:

\[
N\in\{64,128,256,\ldots,16384,\text{Full}\}.
\]

We use deterministic nested subsets, sampled at the problem level before conversion into step-level examples, while keeping the validation set, test set, and number of training epochs fixed. We compare classification and generation SFT on Qwen2.5-0.5B and Qwen2.5-3B, holding the backbone fixed within each comparison, and include ModernBERT-B as an encoder reference. Each configuration below the full-data budget is evaluated with a single run.

\par
\begin{figure}[t!]
    \centering
    \includegraphics[width=1\linewidth]{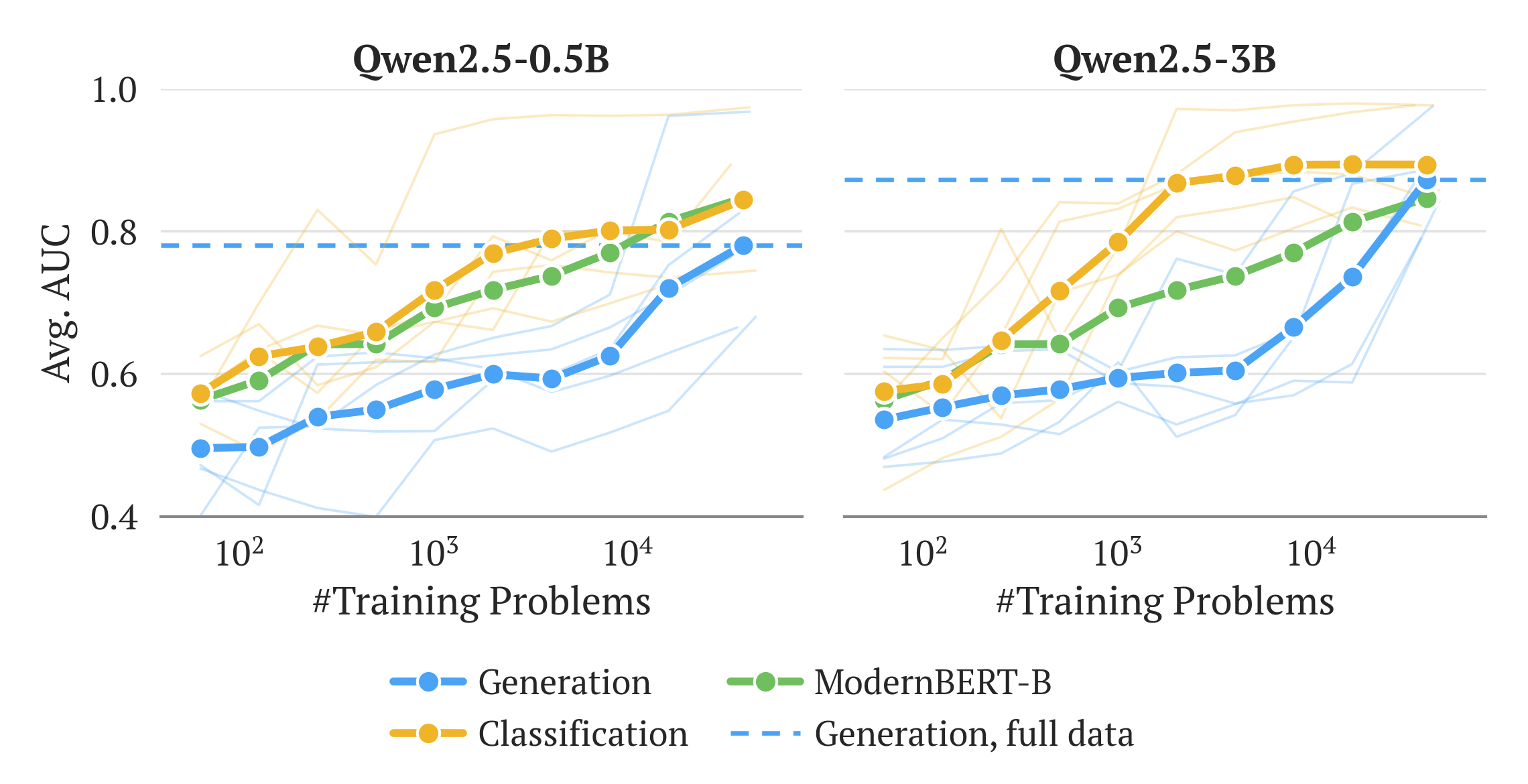}
    \caption{\textbf{Classification is more data-efficient than verdict generation for stepwise verification.} Step-level AUC averaged over the five datasets (faint lines: individual datasets) as the number of annotated training problems increases. On both Qwen2.5-0.5B and Qwen2.5-3B, classification reaches high AUC with far fewer problems than verdict generation; the dashed line marks generation trained on the full set, and ModernBERT-B (0.15B) is the encoder reference.}
    \label{fig:data_scaling}
\end{figure}

 Figure~\ref{fig:data_scaling} reports step-level AUC averaged over the five datasets, with the dashed line marking the corresponding full-data generation baseline. At both model sizes, classification outperforms generation on average at every budget dataset from $1{,}024$ problems onwards. With just $4{,}096$ training problems, classification matches the corresponding full-data generation baseline on average, using only $9$--$12\%$ as many training problems. Generation makes its largest gains at the highest budgets and remains below the 0.15B ModernBERT-B at smaller budgets. These results demonstrate that classification achieves strong verification with substantially fewer annotated problems, an advantage that full-data comparisons alone do not capture.


\subsection{Representation Analysis}
\label{subsec:repr_analysis}

\begin{figure*}[t!]
    \centering
    \includegraphics[width=1\linewidth]{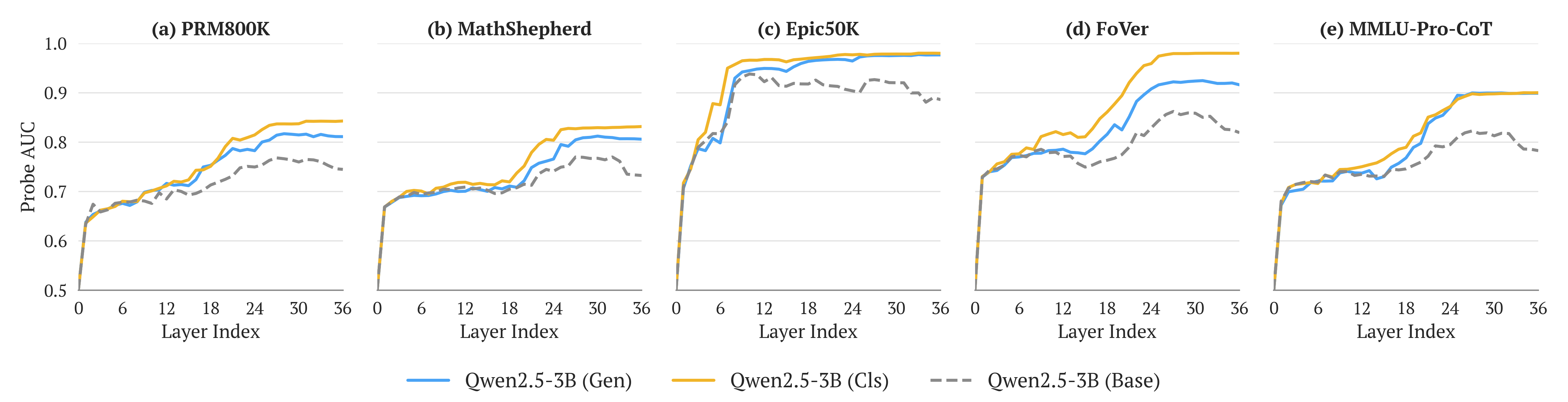}
    \caption{\textbf{Classification SFT makes step-verification labels at least as linearly recoverable as generation SFT in later decoder layers on all five datasets.} Layer-wise probe AUC for the pretrained Qwen2.5-3B backbone, verdict-generation SFT, and classification SFT.}
    \label{fig:probe_layerwise_all}
\end{figure*}

\par We test whether classification SFT makes valid and invalid reasoning steps easier to distinguish from the model's intermediate hidden states. This examines whether its advantage extends beyond the final prediction head to the representations learned within the backbone. We compare pretrained Qwen2.5-3B with its generation-SFT and classification-SFT variants. With each model frozen, we fit a separate logistic-regression probe at each layer to predict $y_i$ from the hidden state at the final non-padding input token~\citep{belinkov2022probing}. Probe AUC measures how readily a linear classifier can recover verification labels from each layer's representations.

Figure~\ref{fig:probe_layerwise_all} suggests that classification's advantage is reflected in intermediate backbone representations, with step-validity labels becoming more linearly recoverable well before the final prediction head.



\subsection{In-Context Learning Baselines}

\begin{table}[t]
    \centering
    \small
    \setlength{\tabcolsep}{3pt}
    \renewcommand{\arraystretch}{1.06}
    \resizebox{\columnwidth}{!}{%
    \begin{tabular}{@{}lrrrrrr@{}}
        \toprule
        \textbf{Model}
        & \textbf{PRM} & \textbf{MS} & \textbf{Epic}
        & \textbf{FoVer} & \textbf{MMLU} & \textbf{Avg.} \\
        \midrule
        \multicolumn{7}{@{}l}{\textit{Qwen2.5-Instruct}} \\
        \quad 0.5B & .491 & .582 & .424 & .485 & .453 & .487 \\
        \quad 1.5B & .528 & .639 & .377 & .575 & .542 & .532 \\
        \quad 3B & .594 & .697 & .494 & .674 & .577 & .607 \\
        \quad 7B & .665 & .708 & .542 & .704 & .696 & .663 \\
        \midrule
        DeepSeek-V4.1-Flash & .779 & .715 & .648 & .773 & .752 & .733 \\
        Gemini-3.8-Flash$^{\dagger}$ & \textbf{.860} & \textbf{.773} & \textbf{.740} & \textbf{.903} & \textbf{.800} & \textbf{.815} \\
        \bottomrule
    \end{tabular}%
    }
    \caption{\textbf{In-context verification improves with scale but stays below a fine-tuned 0.15B encoder on average.} Step-level AUC of in-context verification with three demonstrations per query, retrieved from the training set, and deterministic scoring. Bold marks the best result per column. \textit{Avg.}\ denotes the macro average AUC over five datasets. $^{\dagger}$The Gemini API does not expose token probabilities, so its AUC is computed from binary \texttt{Yes}/\texttt{No} verdicts rather than a continuous score.}
    \label{tab:icl_results}
\end{table}

We test whether LLMs can perform stepwise verification from labelled in-context examples, assessing whether prompting can substitute for task-specific training (Table~\ref{tab:icl_results}). We evaluate Qwen2.5-Instruct from 0.5B to 7B and two API models, DeepSeek-V4.1-Flash~\citep{deepseek2026deepseek} and Gemini-3.8-Flash~\citep{google2026gemini38flash}. Each prompt contains the premise, the hypothesis, and three labelled demonstrations retrieved by embedding similarity~\citep{rubin2022learning, wang2024learning} from the same training pool used for fine-tuning. This keeps the source of supervision fixed, while providing it through in-context examples without parameter updates. We score each step using the \texttt{Yes}/\texttt{No} verdict probabilities. Including the demonstrations' reasoning contexts makes these prompts three to four times as long as the corresponding fine-tuning inputs, requiring a substantial input token budget even in this three-shot setting.

Under this protocol, no prompted model matches the fine-tuned 0.15B ModernBERT-B classifier on average. Qwen2.5 benefits from scale but reaches only $0.663$ average AUC at 7B. The API models perform better, achieving $0.733$ for DeepSeek-V4.1-Flash and $0.815$ for Gemini-3.8-Flash. However, ModernBERT-B achieves $0.847$, exceeding even the strongest prompting baseline by $0.032$ AUC with only 150 million parameters. These results extend the practical case for small classifiers beyond comparisons with generative fine-tuning: task-specific training enables a small verifier to outperform every evaluated prompting baseline on average, despite their access to labelled demonstrations at inference time. Together with the matched-backbone comparisons, they support our central finding that how a model is trained to verify reasoning can matter more than its size.



\section{Discussion and Conclusion}
\label{sec:discussion}

\paragraph{Why classification helps.}
Classification has its largest advantage with smaller backbones and less training data. This advantage persists under verbalizer initialisation, where both objectives start from the same pretrained verdict logits. Generation normalises these logits over the full vocabulary and must suppress non-verdict tokens, whereas classification normalises only over the verdict labels. We conjecture that this additional requirement contributes to generation's poorer performance when capacity or supervision is limited. This interpretation is consistent with prior findings on the sample efficiency of discriminative training~\citep{gu2023don} and the greater gains from scaling generative verifiers~\citep{zhang2025generative}, as well as the disappearance of the gap on Qwen2.5-Math-7B. Our probing results also suggest that classification changes the information recoverable from backbone representations, although probing establishes that information is accessible, not that the model uses it in prediction~\citep{belinkov2022probing}.

\paragraph{Why initialisation matters.}
Classification head initialisation can change which objective performs better. A zero-initialised head transmits no gradient to the backbone on the first backward pass; in our experiments, some seeds also stall well below the performance of other runs (Section~\ref{subsec:head_init}). Initialising from the verdict tokens reduces this variability and, at 3B, reverses the ranking against generation (Tables~\ref{tab:auc_results} and~\ref{tab:head_init}). An unsuitable head initialisation can therefore lead comparisons to understate the effectiveness of classification. We recommend verdict-token initialisation and reporting results across random seeds.

\paragraph{Implications for verifier design.}
Our results support a small classifier trained on step labels as the default for direct stepwise verification, particularly when limited resources must support repeated checks throughout a solution. Domain-specific pretraining also matters: Qwen2.5-Math-1.5B with classification achieves $0.916$ average AUC, outperforming the general-purpose 3B model under either objective. Our comparisons hold the backbone, training instances, prompt, and number of epochs fixed, while using each objective's standard training recipe. They concern direct-verdict prediction; generative verifiers trained on verification rationales can also be data-efficient~\citep{khalifa2025process, lee2025rethinking}. Whether rationale supervision, test-time sampling, or scaling beyond 7B changes the relative advantage of classification remains an open question. Comparing these settings under the same controls would help distinguish the benefits of the generative objective from those of additional supervision and computation.

\par By formulating stepwise verification as NLI, we show that small classifiers can outperform generative verifiers with substantially more parameters and learn effectively from fewer annotated problems. Their advantage extends to the evaluated in-context baselines, where providing labelled demonstrations does not match task-specific training. These findings challenge the assumption that effective verification requires large generative models: how a verifier is trained can matter more than its size. Small classifiers therefore offer a practical approach to checking reasoning when repeated verification must fit within limited data and compute budgets.

\section*{Limitations}
\label{sec:limit}


\paragraph{Annotation requirement.}
Classification makes more efficient use of step-level labels than generation, but does not eliminate the need for supervision. With only a few hundred annotated problems, all models in our data-scaling experiments remain substantially below their full-data AUC (Section~\ref{subsec:data_results}). The benefit is therefore a reduction in annotation requirements, rather than independence from labelled data. For domains without existing step-level annotations, obtaining suitable training data remains a prerequisite for task-specific fine-tuning.

\paragraph{First-error identification.}
Our evaluation stops at the first annotated error, focusing on the earliest point at which a reasoning trajectory becomes invalid. This targets a practical use of verification: identifying a mistake before subsequent reasoning builds on it. Once an error has occurred, a later step may follow from an incorrect premise without being correct relative to the original problem, requiring a distinction between local entailment and overall correctness. Our protocol isolates first-error identification; it does not assess the detection of subsequent errors or recovery from earlier mistakes.

\paragraph{Task complexity.}
Our results demonstrate the effectiveness of small classifiers across mathematical, formal, and multi-domain reasoning tasks of varying difficulty. However, all five datasets contain solutions to self-contained problems, and performance on long formal proofs or research-level mathematics remains untested. Such tasks may involve dependencies beyond the evaluated models' input windows, restricting access to premises needed for verification. This leaves open whether classification retains its advantages as reasoning depth and context requirements increase.





\section*{Ethical Considerations}

\paragraph{Data and licences.}
We use five publicly released datasets and publicly released or API-served models under their respective licences and terms of use (Appendix~\ref{appendix:artifacts}), and release no new dataset. Our use of the datasets follows their intended purpose of training and evaluating step-level verifiers of model-generated reasoning. All five are in English and contain model-generated solutions to mathematical, formal-logic, and subject-knowledge problems, with human or automatically assigned step labels. We did not conduct additional screening for personal or offensive content.

\paragraph{Potential benefits and risks.}
Small classifiers can reduce the computational cost of repeated verification relative to larger generative verifiers, making these checks more feasible in settings with limited resources. However, a verifier may accept an invalid step or reject a valid one. In educational applications, either error could mislead learners if predictions were presented as authoritative feedback. Verifier outputs should therefore support rather than replace human judgement. Errors and inconsistencies in the training annotations, including automatically assigned labels, may also affect the reliability of these predictions.

\paragraph{Use of AI assistants.}
We used AI assistants to support code development and language editing. The authors designed and verified the experiments and analyses and take responsibility for all reported results and claims.

\section*{Acknowledgments}


We gratefully acknowledge the computational resources provided by the Alan Turing Institute and the University of Cambridge High Performance Computing (HPC) Service / CSD3. This material is based on research that is in part supported by DARPA for the SciFy program under agreement number HR00112520301. The U.S. Government is authorized to reproduce and distribute reprints for Governmental purposes notwithstanding any copyright notation thereon. The views and conclusions contained herein are those of the authors and should not be interpreted as necessarily representing the official policies or endorsements, either express or implied, of the DARPA or the U.S. Government.

\bibliography{custom}

\appendix

\section{Literature Review}
\label{appendix:literature}

\par Verification objectives differ in the form of the judgement. Discriminative reward models and PRMs \citep{cobbe2021training, lightman2023let, wang2024math, zhang2025lessons, xia2025evaluating, gao2025llm} produce a classification score, while generative verifiers \citep{zhang2025generative, khalifa2025process, zhao2026genprm, zhong2025dyve, she2025r, xiong2025stepwiser, zhu2026camel} emit a token-level verdict, often after generating verification reasoning. DPO-trained verifiers \citep{hosseini2024v} derive a score implicitly from the divergence between a policy and a reference model, and prompted general-purpose judges \citep{zheng2023judging, kwok2026llm} use no task-specific training.

\par Table~\ref{tab:reasoning_verification} lists the prior studies that compare classification with generation or use encoder verifiers, and marks which factors each holds fixed. GenRM~\citep{zhang2025generative} is the most controlled, comparing a discriminative reward model with direct and rationale-generating verifiers on the same Gemma backbones, but it verifies whole solutions rather than steps. At the step level, ThinkPRM~\citep{khalifa2025process} and \citet{lee2025rethinking} train their generative verifiers on verification rationales, and ThinkPRM uses about a hundredth of the labels of its discriminative baseline; GenPRM~\citep{zhao2026genprm} compares against classification PRMs released with other backbones and data. StepWiser~\citep{xiong2025stepwiser} treats direct-verdict generation as a discriminative judge and contrasts it only with judges that reason first, so classification and direct generation are never separated. Encoders appear as off-the-shelf NLI scorers~\citep{prasad2023receval} or as answer-level verifiers~\citep{li2023making}, not as step classifiers compared with decoders. Our comparison holds the backbone, labelled instances, and prompt fixed, trains the generative verifier to predict the verdict directly, adds encoder classifiers, and varies the annotation budget, so differences between the two objectives cannot be attributed to these factors.

\par Other work is adjacent: training objectives for discriminative scores (contrastive step scoring \citep{khalifa2023grace}, pairwise preference \citep{he2024advancing}, step-progress estimation \citep{setlur2025rewarding}), architectural changes to a discriminative PRM \citep{zhang2026bidirectional}, and the reliability of generated verification under repeated sampling \citep{ye2025uncertainty}. These address the reward-modelling objective or how a verifier is deployed at decoding time rather than the form of the judgement, so we discuss them here but omit them from the table.

\providecommand{\cmark}{$\checkmark$}
\providecommand{\pmark}{($\checkmark$)}

\begin{table*}[t]
    \centering
    \small
    \setlength{\tabcolsep}{5pt}
    \begin{tabular}{@{}lccccccc@{}}
    \toprule
     & \textbf{Step} & \textbf{Cls.\ vs.} & \textbf{Same} & \textbf{Same} & \textbf{Gen.\ without} & \textbf{Encoder} & \textbf{Label budget} \\
    \textbf{Work} & \textbf{level} & \textbf{Gen.} & \textbf{backbone} & \textbf{labels} & \textbf{rationale} & \textbf{classifiers} & \textbf{varied} \\
    \midrule
    GenRM~\citep{zhang2025generative} & -- & \cmark & \cmark & \cmark & \cmark & -- & -- \\
    ThinkPRM~\citep{khalifa2025process} & \cmark & \cmark & \cmark & -- & -- & -- & -- \\
    GenPRM~\citep{zhao2026genprm} & \cmark & \cmark & -- & -- & \cmark & -- & -- \\
    StepWiser~\citep{xiong2025stepwiser} & \cmark & \cmark & \cmark & \cmark & \pmark$^{a}$ & -- & -- \\
    \citet{lee2025rethinking} & \cmark & \cmark & \cmark & -- & -- & -- & \pmark$^{b}$ \\
    ReCEval~\citep{prasad2023receval} & \cmark & -- & -- & -- & -- & \pmark$^{c}$ & -- \\
    DIVERSE~\citep{li2023making} & -- & -- & -- & -- & -- & \cmark & -- \\
    \midrule
    \textbf{Ours} & \cmark & \cmark & \cmark & \cmark & \cmark & \cmark & \cmark \\
    \bottomrule
    \end{tabular}
    \caption{\textbf{Our comparison isolates the training objective for step-level verification.} Each row is a prior study that compares classification with generation or uses encoder verifiers. \textit{Step level}: verifies the validity of individual steps; \textit{Cls.\ vs.\ Gen.}: compares a classifier with a generative verifier; \textit{Same backbone} and \textit{Same labels}: both are trained from the same pretrained model on the same labelled instances; \textit{Gen.\ without rationale}: the generative verifier predicts its verdict directly; \textit{Label budget varied}: the amount of annotated data is varied for both. \cmark: yes; \pmark: partly; --: no. $^{a}$Groups direct-verdict generation with classification as discriminative and contrasts both only with rationale-generating judges. $^{b}$Single-domain versus multi-domain training. $^{c}$Off-the-shelf NLI models, not trained on step labels.}
    \label{tab:reasoning_verification}
\end{table*}

\section{Dataset Construction and Details}
\label{appendix:datasets}

\subsection{Dataset Sources and Annotation Semantics}

\paragraph{Selection rationale.}
\par We select datasets with multi-step reasoning traces, step-level annotations, and minority-class supervision at a scale suitable for full-parameter classification fine-tuning. All five datasets are in English. Process-supervision corpora are a natural starting point because they attach feedback to intermediate reasoning states rather than outcomes alone \citep{lee2025evaluating,zheng2025survey}. Our mathematical core (PRM800K, MathShepherd, and Epic50K) holds the task domain approximately fixed while varying supervision among human correctness judgements, rollout-based answer reachability, and contribution-based automatic annotations. Section~\ref{subsec:models_datasets} discusses how these sources differ in what a label measures. FoVer and MMLU-Pro-CoT extend the evaluation to formally grounded and multi-domain reasoning, respectively, testing whether findings persist across annotation regimes and beyond mathematics.

\begin{itemize}
    \item \textbf{PRM800K.} PRM800K contains model-generated solutions to competition-level MATH problems with human judgements of whether each intermediate step is correct and reasonable \citep{lightman2023let}. It is the closest of the five datasets to direct local-validity supervision; we normalise its released labels to entailment and non-entailment.
    \item \textbf{MathShepherd.} MathShepherd provides automatic process supervision for GSM8K and MATH solutions by sampling continuations from intermediate states and checking their final answers \citep{wang2024math}.
    \item \textbf{Epic50K.} Epic50K provides contribution-based automatic annotations for mathematical reasoning steps \citep{sun2025efficient}. From each released record, we reconstruct prefix--candidate trajectories, treating the accepted prefix as valid and using the candidate rating as the next-step label.
    \item \textbf{FoVer.} FoVer contains LLM-generated formal-logic and formal-proof traces labelled using formal verification tools, including Z3 and Isabelle \citep{kamoi2026efficient}. We use the balanced Qwen2.5-7B subset combining both sources.
    \item \textbf{MMLU-Pro-CoT.} MMLU-Pro-CoT is the multi-domain synthetic reasoning corpus released with VersaPRM \citep{zeng2025versaprm}. We use its step annotations across mathematical, scientific, engineering, social-science, and other knowledge-intensive tasks.
\end{itemize}

\subsection{Preprocessing and Experimental Formatting}

\paragraph{Verification instance construction.}
\par
\begin{figure}[htbp]
    \centering
    \includegraphics[width=1\linewidth]{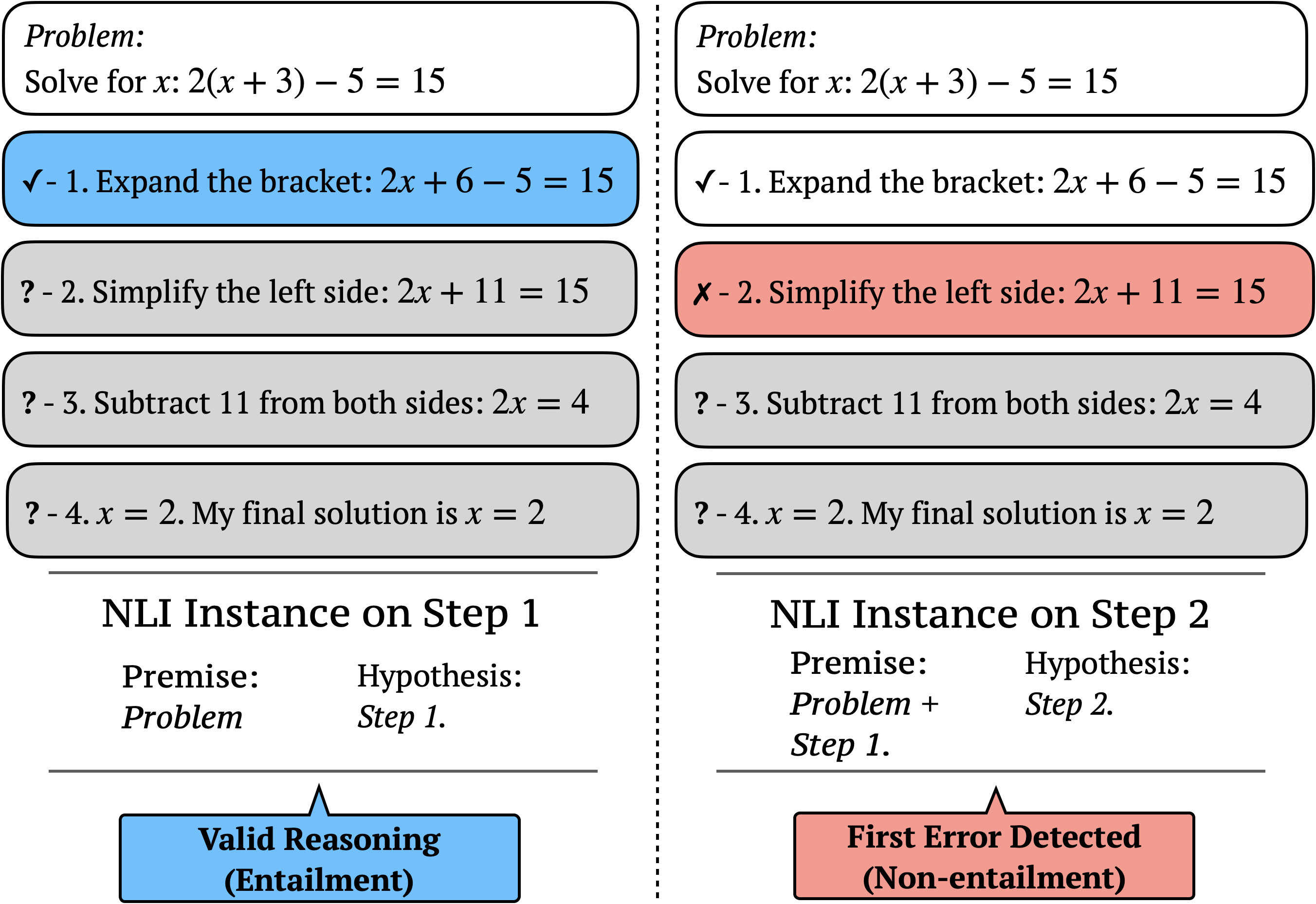}
    \caption{\textbf{Reformulating reasoning verification as first-error localisation.} We cast verification as identifying the first step where reasoning ceases to be logically valid. This is naturally formulated as sequential natural language inference, where the validated reasoning prefix is treated as the premise and the current reasoning step as the hypothesis. }
    \label{fig:nli_example}
\end{figure}

 We first normalise each dataset into a shared schema containing a problem, ordered reasoning steps, and aligned labels. After constructing the data splits, we expand each trajectory into the local NLI instances defined in Section~\ref{subsec:task_form}, stopping after the first annotated error so that every hypothesis is evaluated against a prefix not already invalidated (Figure~\ref{fig:nli_example}).

\par A trajectory whose first error occurs at position $k$ contributes $k-1$ entailment instances but only one non-entailment instance, while a correct trajectory contributes only entailment instances. This imbalance is why the training split is undersampled within each step position (Section~\ref{subsec:models_datasets}), which preserves the position distribution.

\paragraph{In-context demonstration construction.}
\par The in-context models receive three demonstrations drawn only from the training partition. The retrieval pool is the full training set used for fine-tuning. When a prompt exceeds the 4,096-token context of the Qwen2.5-Instruct runs, demonstrations are dropped, oldest first, so the query step is never truncated. Demonstrations are retrieved by cosine similarity over \texttt{all-MiniLM-L6-v2} embeddings of the candidate step and final 120 premise words \citep{reimers2019sentence, rubin2022learning, wang2024learning}; each prompt includes both labels and prepends the demonstrations to the base prompt below. The prompt requests a direct verdict without a rationale.

\paragraph{Supervised training formats.}
\par All supervised conditions use the same verification instances and binary supervision. Encoders receive paired premise--hypothesis inputs; generation SFT renders the same content with the template below and learns a complete \texttt{Yes}/\texttt{No} response; decoder classifiers predict the label from the final non-padding representation. We provide neither demonstrations nor verification rationales, preventing retrieval, additional supervision, and iterative generation from confounding the comparison. Thus, all conditions share input semantics and supervision, while the two decoder paradigms additionally share the backbone and differ primarily in whether the verdict is learned through next-token generation or sequence classification.

\begin{promptbox}{Base Verification Prompt}
    You are a strict mathematics tutor. Verify the logic of the following step.
    
    \#\#\# Context: \\
    \{problem\_description\} \\
    Step 1: \{step\_1\} \\
    Step 2: \{step\_2\} \\
    ...
    
    \#\#\# Step to Verify: \\
    \{current\_step\_to\_verify\}
    
    \#\#\# Verification: \\
    Is this step correct? Answer with `Yes' or `No'.
\end{promptbox}

\subsection{Dataset Statistics and Splits}

\par We construct deterministic train, validation, and test partitions using the dataset seed before expanding trajectories into NLI instances. Partitions are split by question, so all trajectories for one question, and all candidate paths reconstructed from one source record, fall in the same partition. For PRM800K and MathShepherd, after removing test questions from training (Section~\ref{subsec:models_datasets}), we hold out 1\% of the remaining training questions for validation. Epic50K, FoVer, and MMLU-Pro-CoT each provide a single source partition: we reserve 10\% of questions for testing and, of the remainder, 5\% for validation on Epic50K and FoVer and 2\% on MMLU-Pro-CoT.

\par To train on comparable amounts of data across datasets, we cap the number of training problems, each a labelled solution trajectory, at 46,000 of 468,066 for PRM800K, 37,000 of 190,607 for MathShepherd, and 38,000 of 74,079 for MMLU-Pro-CoT, sampled with the dataset seed. Epic50K (42,697) and FoVer (34,302) are used in full. Validation and test partitions are never subsampled.

\par Table~\ref{tab:dataset_statistics} reports the number of premise--hypothesis instances after dataset normalisation, first-error truncation, and NLI conversion; these counts therefore differ from the number of original reasoning trajectories. Training counts are after merging duplicates and undersampling; the natural error ratio is the proportion of non-entailment instances in the unbalanced test split. We additionally report token-length statistics for the complete test-set NLI inputs before model-specific truncation. Lengths are measured using the BERT tokenizer, and the overflow columns show the proportion exceeding 512 and 2,048 tokens.

\begin{table*}[htbp]
    \centering
    \begingroup
    \small
    \resizebox{\textwidth}{!}{%
    \begin{tabular}{l rrr c rrrr crr}
        \toprule
        \multirow{2}{*}{\textbf{Dataset}} & \multicolumn{3}{c}{\textbf{Instances}} & \textbf{Natural Error} & \multicolumn{4}{c}{\textbf{Token Length Statistics (Test)}} & \multicolumn{2}{c}{\textbf{Overflow Ratio (\%)}} \\
        \cmidrule(lr){2-4} \cmidrule(lr){6-9} \cmidrule(lr){10-11}
         & \textbf{Train} & \textbf{Val} & \textbf{Test} & \textbf{Ratio (\%)} & \textbf{Mean $\pm$ Std} & \textbf{Med.} & \textbf{P90} & \textbf{Max} & \textbf{$>512$} & \textbf{$>2048$} \\
        \midrule
        \textbf{PRM800K}       &  39,742 & 19,575 & 70,548 & 8.3\%  & $231.8 \pm 152.5$ & 191 & 438 & 1,262 & 6.1\% & 0.0\% \\
        \textbf{MathShepherd}  &  43,667 &  8,570 & 80,883 & 17.5\%  & $185.3 \pm 131.9$ & 146 & 341 & 1,580 & 3.0\% & 0.0\% \\
        \textbf{Epic50K}       &  58,151 & 15,982 & 37,763 & 6.0\%  & $651.5 \pm 861.4$ & 315 & 1,643 & 5,399 & 33.0\% & 7.8\% \\
        \textbf{FoVer}         &  44,399 &  5,464 & 10,806 & 24.0\%  & $304.9 \pm 148.5$ & 266 & 529 & 936 & 11.6\% & 0.0\% \\
        \textbf{MMLU-Pro-CoT}  &  42,612 &  8,727 & 49,494 & 9.6\%  & $343.9 \pm 228.3$ & 286 & 602 & 2,351 & 15.6\% & 0.1\% \\
        \bottomrule
    \end{tabular}%
    }
    \endgroup

    \caption{\textbf{Dataset statistics and token lengths.} Number of premise--hypothesis instances per split, the percentage of non-entailment instances in the test split, and token-length statistics of the test inputs. Training splits are balanced between entailment and non-entailment by local undersampling, whereas validation and test splits keep their natural distribution. PRM800K, MathShepherd, and MMLU-Pro-CoT are capped at 37k--46k training problems before instance construction; Epic50K and FoVer use all problems. Token length is measured with the \texttt{bert-base-cased} tokenizer on the premise--hypothesis pair.}
    \label{tab:dataset_statistics}
\end{table*}

\section{Implementation Details}
\label{appendix:impl_details}

\subsection{Models}
\label{subsec:models}

\begin{table}[t]
    \centering
    \begingroup
    \small
    \setlength{\tabcolsep}{3.5pt}
    \renewcommand{\arraystretch}{1.05}

    \begin{tabular}{@{}lrrr@{}}
        \toprule
        \textbf{Model}
        & \textbf{Params.}
        & \multicolumn{2}{c}{\textbf{Max Length}} \\
        \cmidrule(lr){3-4}
        & & \textbf{Native} & \textbf{Used} \\
        \midrule

        \multicolumn{4}{@{}l}{\textit{Encoders}} \\
        BERT-B            & 0.110B & 512     & 512   \\
        RoBERTa-B         & 0.125B & 512     & 512   \\
        ModernBERT-B      & 0.149B & 8,192   & 2,048 \\
        DeBERTa-B         & 0.184B & 512     & 512   \\
        BERT-L            & 0.340B & 512     & 512   \\
        RoBERTa-L         & 0.355B & 512     & 512   \\
        ModernBERT-L      & 0.395B & 8,192   & 2,048 \\
        DeBERTa-L         & 0.435B & 512     & 512   \\

        \midrule

        \multicolumn{4}{@{}l}{\textit{Decoders}} \\
        Qwen2.5-0.5B      & 0.49B  & 32,768  & 2,048 \\
        Qwen2.5-1.5B      & 1.54B  & 32,768  & 2,048 \\
        Qwen2.5-Math-1.5B & 1.54B  & 4,096  & 2,048 \\
        Qwen2.5-3B        & 3.09B  & 32,768  & 2,048 \\
        Llama-3.2-3B      & 3.21B  & 131,072 & 2,048 \\
        Gemma-3-4B        & 3.88B  & 131,072 & 2,048 \\
        Qwen2.5-7B        & 7.61B  & 131,072 & 2,048 \\
        Qwen2.5-Math-7B   & 7.61B  & 4,096 & 2,048 \\

        \bottomrule
    \end{tabular}
    \endgroup

    \caption{\textbf{Models evaluated in this work.} Parameters denote the total parameter count, while model names retain their developers' nominal size labels. DeBERTa denotes DeBERTa-v3, and Gemma-3-4B is its text-only language model. Native denotes the maximum sequence length supported by the original checkpoint, and Used denotes the maximum sequence length used during fine-tuning and evaluation.}
    \label{tab:models}
\end{table}

\par Table~\ref{tab:models} summarises the evaluated backbones, parameter counts and context limits. The encoder suite covers four architectural families at base and large scales, allowing comparisons across both architecture and capacity. The decoder suite centres on Qwen2.5 to support scaling comparisons and matched generation--classification experiments within a common model family; Qwen2.5-Math tests domain-specialised pre-training, while Llama-3.2-3B and Gemma-3-4B provide a cross-family comparison.

\paragraph{Context-length control.}
\par We use 512 tokens for BERT, RoBERTa, and DeBERTa, and 2,048 tokens for ModernBERT and every decoder condition, including decoder classification. Model inputs exceeding these limits are truncated from the left under the same policy during training and evaluation, preferentially removing the earliest premise tokens while retaining the candidate step and recent reasoning context. Table~\ref{tab:dataset_statistics} reports overflow before this truncation.

\paragraph{Training precision.}
\par Every trained model keeps FP32 master parameters and uses BF16 autocast, except DeBERTa, which trains in full FP32. Distributed sharding changes memory placement but not the model objective or input construction.

\subsection{Training Configuration}

\par We use full-parameter fine-tuning with 8-bit AdamW \citep{dettmers2022optimizers}, cosine learning-rate decay, a weight decay of 0.01, and gradient clipping at 1.0. All supervised experiments are run with seeds 1, 42, and 3407 on fixed data partitions. Each seed is launched in a fresh process, and the global random state is set before model construction so that parameter initialisation, data ordering, and training stochasticity vary without changing the underlying split. Encoder learning rates lie between $1\times10^{-5}$ and $3\times10^{-5}$; we use commonly adopted fine-tuning defaults for each encoder. Decoder generation SFT uses $2\times10^{-5}$ and decoder classification uses $1\times10^{-5}$. We train encoders for at most 5 epochs and decoders for 1 epoch. We use a global batch size of 32. 

\par Encoder configurations follow standard full-parameter fine-tuning ranges rather than a single setting imposed across architectures. Encoders use a warmup ratio of 0.1, reduced to 0.02 on MathShepherd, and are evaluated periodically on the validation partition; the checkpoint with the highest validation AUC is retained, with early stopping based on the same metric. Decoders are trained for one epoch and keep the final checkpoint.

\par Generation and classification models are trained on the same verification instances and data partitions. Matched decoder conditions additionally share the backbone, context limit, cosine schedule, prescribed number of epochs, and final-checkpoint rule, isolating the principal experimental difference to whether the verdict is learned through next-token generation or a classification head. Generation SFT uses sequence packing, whereas classification processes verification instances independently; optimiser-step counts are therefore not directly comparable across the two objectives. We instead control data exposure by prescribing the number of epochs consistently within each paradigm. No test-set results are used for hyperparameter or checkpoint selection.

\subsection{Evaluation Protocol}

\par All supervised models produce a continuous non-entailment score for each verification instance. Encoder and decoder classifiers obtain this score by applying a two-class softmax to their classification logits. For generation SFT, we teacher-force the exact single-token \texttt{Yes} and \texttt{No} verbalizers, retrieve their conditional log-probabilities with \texttt{vLLM}, and normalise over the two alternatives. Both verbalizer-extended sequences are evaluated in one batched call with prefix caching, which reuses their shared prompt and avoids top-$k$ approximation or free-form output parsing. We verify that each verbalizer contributes exactly one token under the model-specific chat template and tokenizer.

\par The same test instances are scored independently by every supervised condition. Step-level AUC uses the continuous scores directly and is therefore threshold-free. Test labels are never used for checkpoint selection or hyperparameter choice.

\par We evaluate each seed separately and report the arithmetic mean and sample standard deviation across seeds. Because generation SFT and decoder classification use different readout mechanisms, their raw optimiser-step counts and inference calls are not treated as matched units; comparisons instead use identical test instances and the common non-entailment scoring interface described above.

\subsection{Software and Hardware}

\par All experiments use Python 3.12, PyTorch 2.11 with CUDA 12.8, Transformers 5.5.3, Datasets 4.0.0, and Accelerate 1.11.0. Generation SFT is run with LlamaFactory 0.9.5 \citep{zheng2024llamafactory}; fine-tuned generation and in-context models are scored with vLLM 0.25.1, and decoders use FlashAttention 2.8.3. Retrieval embeddings are computed with Sentence-Transformers 5.6.1 and AUC with scikit-learn 1.9.0. 

\par All models are trained and evaluated on NVIDIA A100-PCIE-40GB GPUs. Table~\ref{tab:compute} gives the number of GPUs and the GPU-hours needed to train one model on one dataset with one seed; each entry of Table~\ref{tab:auc_results} is three such runs per dataset. Training all models in Table~\ref{tab:auc_results} takes about 950 GPU-hours in total.

\begin{table}[htbp]
    \centering
    \small
    \setlength{\tabcolsep}{4pt}
    \begin{tabular}{@{}lcc@{}}
        \toprule
        \textbf{Model} & \textbf{GPUs} & \textbf{GPU-hours} \\
        \midrule
        Encoders, base & 1 & 0.2--0.5 \\
        Encoders, large & 1 & 0.3--1.1 \\
        \midrule
        Decoders 0.5B & 1 & 0.3--0.4 \\
        Decoders 1.5B & 1 & 0.5--0.8 \\
        Decoders 3B--4B, Gen. & 2 & 1.7--2.6 \\
        Decoders 3B--4B, Cls. & 2 & 2.5--3.3 \\
        Decoders 7B, Gen. & 4 & 7.8 \\
        Decoders 7B, Cls. & 4 & 13.0 \\
        \bottomrule
    \end{tabular}
    \caption{\textbf{Training cost grows from around one GPU-hour for encoders to thirteen for 7B decoder classifiers.} GPUs used per training run and GPU-hours per run on NVIDIA A100-PCIE-40GB, averaged over the five datasets; ranges span the models in each group.}
    \label{tab:compute}
\end{table}


\section{Layer-wise Probing Protocol}
\label{appendix:repr_setting}

\par We analyse pretrained, generation-fine-tuned, and classification-fine-tuned Qwen2.5-3B checkpoints across all five evaluation datasets, using the seed-1 checkpoint of each fine-tuned model. For each dataset, the backbone is frozen and the three model states use the same deterministically sampled probe instances.

\par We extract the hidden state at the final non-padding token from the embedding output and every transformer block, holding the readout position fixed across model states. All three model states read the same rendered prompt as the fine-tuned decoders (Appendix~\ref{appendix:datasets}). We use at most 5,000 training instances and 5,000 validation instances. At each layer, we fit an independent standardised logistic-regression probe with L-BFGS, balanced class weights, tolerance $10^{-3}$, and at most 1,000 iterations. The inverse regularisation strength $C\in\{10^{-4},10^{-3},10^{-2},10^{-1},1\}$ is selected by AUC on a stratified 20\% hold-out of the probe training instances, and the probe is then refit on all of them. The backbone remains frozen, and the same procedure is applied to every layer and model state.

\section{Artifact Licences}
\label{appendix:artifacts}

\par Table~\ref{tab:artifact_licences} lists the datasets and models we use with the licence under which each is distributed. We use every artifact for research only and release no new dataset.

\begin{table}[htbp]
    \centering
    \footnotesize
    \setlength{\tabcolsep}{3pt}
    \begin{tabularx}{\columnwidth}{@{}Xl@{}}
        \toprule
        \textbf{Artifact} & \textbf{Licence} \\
        \midrule
        \multicolumn{2}{@{}l}{\textit{Datasets}} \\
        PRM800K~\citep{lightman2023let} & MIT \\
        MathShepherd~\citep{wang2024math} & None stated \\
        Epic50K~\citep{sun2025efficient} & Apache-2.0 \\
        FoVer~\citep{kamoi2026efficient} & CC BY 4.0 \\
        MMLU-Pro-CoT~\citep{zeng2025versaprm} & MIT \\
        \midrule
        \multicolumn{2}{@{}l}{\textit{Models}} \\
        BERT~\citep{devlin2019bert} & Apache-2.0 \\
        RoBERTa~\citep{liu2019roberta} & MIT \\
        DeBERTa~\citep{he2023debertav3} & MIT \\
        ModernBERT~\citep{warner2025smarter} & Apache-2.0 \\
        Qwen2.5 (0.5B, 1.5B, 7B) & Apache-2.0 \\
        Qwen2.5 (3B)~\citep{yang2024qwen2} & Qwen Research \\
        Qwen2.5-Math~\citep{yang2024qwen2math} & Apache-2.0 \\
        Llama-3.2-3B~\citep{grattafiori2024llama} & Llama 3.2 \\
        Gemma-3-4B~\citep{team2025gemma} & Gemma Terms \\
        Gemini-3.8-Flash~\citep{google2026gemini38flash} & API Terms \\
        DeepSeek-V4.1-Flash~\citep{deepseek2026deepseek} & API Terms \\
        all-MiniLM-L6-v2~\citep{reimers2019sentence} & Apache-2.0 \\
        \bottomrule
    \end{tabularx}
    \caption{\textbf{Every artifact we use is publicly released and used here for research.} Licences are those stated by the original release, or by the model card for models. Qwen2.5 licences apply to both the base and the Instruct checkpoints; Llama-3.2 is under the Llama 3.2 Community License, Gemma under the Gemma Terms of Use, and Gemini-3.8-Flash and DeepSeek-V4.1-Flash are accessed through their providers' APIs under the respective API terms of service. MathShepherd is distributed publicly by its authors without a stated licence.}
    \label{tab:artifact_licences}
\end{table}

\end{document}